# Reasoning about Probabilities in Dynamic Systems using Goal Regression[*]

**Vaishak Belle** and **Hector J. Levesque**
Dept. of Computer Science
University of Toronto
Toronto, Ontario M5S 3H5, Canada
{vaishak, hector}@cs.toronto.edu

## Abstract

Reasoning about degrees of belief in uncertain dynamic worlds is fundamental to many applications, such as robotics and planning, where actions modify state properties and sensors provide measurements, both of which are prone to noise. With the exception of limited cases such as Gaussian processes over linear phenomena, belief state evolution can be complex and hard to reason with in a general way. This paper proposes a framework with new results that allows the reduction of subjective probabilities after sensing and acting, both in discrete and continuous domains, to questions about the initial state only. We build on an expressive probabilistic first-order logical account by Bacchus, Halpern and Levesque, resulting in a methodology that, in principle, can be coupled with a variety of existing inference solutions.

## 1 INTRODUCTION

Reasoning about *degrees of belief* in uncertain dynamic worlds is fundamental to many applications, such as robotics and planning, where actions modify state properties and sensors provide measurements, both of which are prone to noise. However, there seem to be two disparate paradigms to address this concern, both of which have their limitations. At one extreme, there are logical formalisms, such as the *situation calculus* (McCarthy and Hayes, 1969; Reiter, 2001), which allows us to express strict uncertainty, and exploits regularities in the effects actions have on propositions to describe physical laws compactly. Probabilistic sensor fusion, however, has received less attention here. At the other extreme, revising beliefs after noisy observations over rich error profiles is effortlessly addressed using *probabilistic techniques* such as Kalman filtering and Dynamic Bayesian Networks (Dean and Kanazawa, 1989; Dean and Wellman, 1991). However, in these frameworks, a complete specification of the dependencies between variables is taken as given, making it difficult to deal with other forms of incomplete knowledge as well as complex actions that shift dependencies between variables in nontrivial ways.

An influential but nevertheless simple proposal by Bacchus, Halpern and Levesque (1999), BHL henceforth, was among the first to merge these broad areas in a general way. Their specification is widely applicable because it is not constrained to particular structural assumptions. In a nutshell, they extend the situation calculus language with a provision for specifying the degrees of belief in formulas in the initial state, closely fashioned after intuitions on incorporating probability in modal logics (Halpern, 1990; Fagin and Halpern, 1994). This then allows incomplete and partial specifications, which might be compatible with one or very many initial distributions and sets of independence assumptions, with beliefs following at a corresponding level of specificity. Moreover, together with a rich action theory, the model not only exhibits Bayesian conditioning (Pearl, 1988) (which, then, captures special cases such as Kalman filtering), but also allows flexibility in the ways dependencies and distributions may change over actions.

What is left open, however, is the following computational concern: how do we effectively reason about degrees of belief in the framework? That is, while changing degrees of belief do indeed emerge as *logical entailments* of the given action theory, no procedure is given for computing these entailments. On closer examination, in fact, this is a two-part question:

(i) How do we effectively reason about beliefs in a particular state?

(ii) How do we effectively reason about belief state evolution and belief change?

In the simplest case, part (i) puts aside acting and sensing, and considers reasoning about the *initial* state only, which is then the classical problem of (first-order) probabilistic inference. We do not attempt to do a full survey here, but this has received a lot of attention, often under reasonable assumptions such as the ability to factorize domains (Poole, 2003; Gogate and Domingos, 2010).

[*]We thank the reviewers for very helpful comments. Financial support from the Natural Science and Engineering Research Council of Canada made this research possible.

This paper is about part (ii). Addressing this concern has a critical bearing on the assumptions made about the domain for tractability purposes. For example, if the initial state supports a decomposed representation of the distribution, can we expect the same after actions? In the exception of very limited cases such as Kalman filtering that harness the conjugate property of Gaussian processes, the situation is discouraging. In fact, even in the slightly more general case of Dynamic Bayesian Networks, which are in essence atomic propositions, if one were to assume that state variables are independent at time 0, they can become fully correlated after a few steps (Dean and Kanazawa, 1988; Boyen and Koller, 1998; Hajishirzi and Amir, 2010). Dealing with complex actions, incomplete specifications and mixed representations, therefore, is significantly more involved.

In this paper, we propose a new alternative to infer degrees of belief in the presence of a rich theory of actions, closely related to *goal regression* (Waldinger, 1977; Reiter, 2001). The procedure is general, not requiring (but allowing) structural constraints about the domain, nor imposing (but allowing) limitations to the family of actions. Regression derives a mathematical formula, using term and formula *substitution* only, that relates belief after a sequence of actions and observations, even when they are noisy, to beliefs about the initial state. That is, among other things, if the initial state supports efficient factorizations, regression will maintain this advantage no matter how actions affect the dependencies between state variables over time. Going further, the formalism will work seamlessly with discrete probability distributions, probability densities, and perhaps most significantly, with difficult combinations of the two. (See Example 9.3 in Section 4 below.)

To see a simple example of what goal regression does, imagine a robot facing a wall and at a certain distance $h$ to it, as in Figure 1. The robot might initially believe $h$ to be drawn from a uniform distribution on $[2, 12]$. Assume the robot moves away by 2 units and is now interested in the belief about $h \leq 5$. Regression would tell the robot that this is equivalent to its initial beliefs about $h \leq 3$ which here would lead to a value of .1. To see a nontrivial example, imagine now the robot is also equipped with a sonar unit aimed at the wall, that adds Gaussian noise with mean $\mu$ and variance $\sigma^2$. After moving away by 2 units, if the sonar were now to provide a reading of 8, then regression would derive that belief about $h \leq 5$ is equivalent to

$$\frac{1}{\gamma} \int_2^3 .1 \times \mathcal{N}(6 - x; \mu, \sigma^2) \, dx.$$

where $\gamma$ is the normalization factor. Essentially, the posterior belief about $h \leq 5$ is reformulated as the product of the prior belief about $h \leq 3$ and the likelihood of $h \leq 3$ given an observation of 6. (That is, observing 8 after moving away by 2 units is related here to observing 6 initially.)

We believe the broader contributions of this line of work are two-fold. On the one hand, as we show later, simple cases of belief state evolution, as applicable to conjugate distributions for example, are special cases of regression's backward chaining procedure. Thus, regression could serve as a *formal basis* to study probabilistic belief change wrt limited forms of actions. On the other hand, our contribution can be viewed as a methodology for combining actions with recent advances in probabilistic inference, because reasoning about actions reduces to reasoning about the initial state.

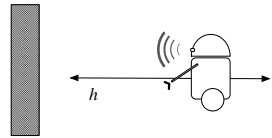

Figure 1: Robot moving towards a wall.

We now describe the structure of the paper. Before moving on, we note that although the original BHL account is only suitable for discrete domains (as they assume values are taken from countable sets), in a companion paper (Belle and Levesque, 2013a) we show that the account can be generalized to domains with both discrete and continuous variables with minimal additions. In the preliminaries section, we cover the situation calculus, recap BHL and go over the essentials of its continuous extension. We then present regression for discrete domains, followed by regression for general domains. We end with related and future work. For this version of the paper, we allow noisy sensors but assume deterministic (noise-free) physical actions. Noisy actions are left for an extended version.

## 2 BACKGROUND

The language $\mathcal{L}$ of the situation calculus (Reiter, 2001) is a many-sorted dialect of predicate calculus, with sorts for *physical actions*, *sensing actions, situations* and *objects* (including the set of reals $\mathbb{R}$ as a subsort). A situation represents a history as a sequence of actions. A set of initial situations correspond to the ways the world might be initially. Successor situations are the result of doing actions, where the term $do(a, s)$ denotes the unique situation obtained on doing $a$ in $s$. The term $do(\alpha, s)$, where $\alpha$ is the sequence $[a_1, \ldots, a_n]$, abbreviates $do(a_n, do(\ldots, do(a_1, s) \ldots))$. For example, $do([grasp(o_1), repair(o_1)], s)$ represents the situation obtained after grasping and repairing object $o_1$ starting from $s$. Initial situations are those without a predecessor:

$$Init(s) \doteq \neg \exists a, s'. \; s = do(a, s').$$

We let the constant $S_0$ denote the actual initial situation, and we use the variable $\iota$ to range over initial situations only. $\mathcal{L}$ also includes functions whose values vary from situation to situation, called *fluents*, whose last argument is a situation.

We follow two notational conventions. We often suppress the situation argument in a formula $\phi$, or use a distinguished variable *now*. Either way, $\phi[t]$ is used to denote the formula with that variable replaced by $t$, e.g. both $(f < 12)[s]$ and $(f(now) < 12)[s]$ mean $f(s) < 12$. We also use conditional *if-then-else* expressions in formulas throughout. We write $f = $ IF $\phi$ THEN $t_1$ ELSE $t_2$ to mean $[\phi \wedge f = t_1] \vee [\neg \phi \wedge f = t_2]$. In case quantifiers appear inside the *if*-condition, we take some liberties with notation and the scope of variables in that we write $f = $ IF $\exists x. \phi$ THEN $t_1$ ELSE $t_2$ to mean $\exists x \, [\phi \wedge f = t_1] \vee [(f = t_2) \wedge \neg \exists x. \phi]$.

**Basic action theory**

Following (Reiter, 2001), we model dynamic domains in $\mathcal{L}$ by means of a *basic action theory* $\mathcal{D}$, which consists of domain-independent *foundational* axioms, *unique name* axioms for actions (see (Reiter, 2001)), and (1) axioms $\mathcal{D}_0$ that describe what is true in the initial states, including $S_0$;[1] (2) precondition axioms[2] of the form $Poss(A(\vec{x}), s) \equiv \Pi_A(\vec{x}, s)$ describing executability conditions using a special fluent *Poss*; and (3) successor state axioms of the following form stipulating how fluents change:

$$f(do(a,s)) = u \equiv \exists \vec{z_1}[a = A_1(\vec{z_1}) \wedge u = e_1(\vec{z_1}, s)] \vee \ldots \\ \exists \vec{z_k}[a = A_k(\vec{z_k}) \wedge u = e_k(\vec{z_k}, s)] \vee \\ \bigwedge \neg \exists \vec{z_i}[a = A_i(\vec{z_i})] \wedge u = f(s). \quad (1)$$

where $e_i(\vec{z_i}, s)$ is any expression whose only free variables are $\vec{z_i}$ and $s$. For example, consider the action $fwd(z)$ of moving precisely $z$ units towards or away from the wall, but the motion stops when the wall is reached:

$$h(do(a,s)) = u \equiv \exists z[a = fwd(z) \wedge u = \max(0, h(s) - z)] \vee \\ \neg \exists z[a = fwd(z)] \wedge u = h(s). \quad (2)$$

This sentence states that $fwd(z)$ is the only action affecting fluent $h$, in effect incorporating a solution to the frame problem (Reiter, 2001). Given an action theory, an agent reasons about actions by means of entailments of $\mathcal{D}$.[3]

**Likelihood and belief**

The BHL model of belief builds on a treatment of *knowledge* in $\mathcal{L}$ (Scherl and Levesque, 2003). Here we present a simpler variant based on two distinguished fluents $l$ and $p$.

The term $l(a, s)$ is intended to denote the likelihood of action $a$ in situation $s$. For example, suppose $sonar(z)$ is the action of reading the value $z$ from a sonar that measures the distance to the wall, $h$. We might assume that this action is characterized by a simple discrete error model (continuous error models are considered later):

$$l(sonar(z), s) = \text{IF } |h(s) - z| \leq 1 \text{ THEN } 1/3 \text{ ELSE } 0 \quad (3)$$

which stipulates that the difference between a reading of $z$ and the true value $h$ is either $\{0, -1, 1\}$ with probability $1/3$, assuming that $h$ and $z$ take integer values. In general, the action theory $\mathcal{D}$ is assumed to contain for each sensor $sense_i(\vec{x})$ that measures a fluent $f$, an axiom of the form:

$$l(sense_i(\vec{x}), s) = Err_i(\vec{x}, f(s)),$$

where $Err_i(u_1, u_2)$ is some expression with only two free variables $u_1$ and $u_2$, both numeric.[4] (Noise-free physical actions are given a likelihood of 1.)

Next, the $p$ fluent determines a (subjective) probability distribution on situations. The term $p(s', s)$ denotes the relative *weight* accorded to situation $s'$ when the agent happens to be in situation $s$, as in modal probability logics (Fagin and Halpern, 1994). Now, the task of the modeler is to specify the initial properties of $p$ as part of $\mathcal{D}_0$ using $\iota$ and $S_0$, e.g.:

$$p(\iota, S_0) = \text{ IF } h(\iota) \in \{2, \ldots, 11\} \text{ THEN } .1 \text{ ELSE } 0 \quad (4)$$

says that $h$ is drawn from a uniform distribution. The following nonnegative constraint is also included in $\mathcal{D}_0$:

$$\forall \iota, s. \, p(s, \iota) \geq 0 \wedge (p(s, \iota) > 0 \supset Init(s)) \quad \text{(P1)}$$

Then, by means of a remarkably simple successor state axiom for $p$, (P2) below, the formal specification is complete.

$$p(s', do(a, s)) = \\ \quad \text{IF } \exists s''. \, s' = do(a, s'') \wedge Poss(a, s'') \\ \quad \quad \text{THEN } p(s'', s) \times l(a, s'') \\ \quad \text{ELSE } 0 \quad \text{(P2)}$$

In particular, the *degree of belief* in a formula $\phi$ can be accounted for in terms of an abbreviation:[5]

$$Bel(\phi, s) \doteq \frac{1}{\gamma} \sum_{\{s': \phi[s']\}} p(s', s) \quad \text{(B)}$$

where $\gamma$, the normalization factor, is understood throughout as the same expression as the numerator but with $\phi$ replaced by *true*, e.g. here $\gamma$ is $\sum_{s'} p(s', s)$. So, as in probability logics, belief is simply the total weight of worlds satisfying $\phi$. But the novelty here is that in a dynamical setting, belief change via (B) is identical to Bayesian conditioning:

---

[1] Note that $\mathcal{D}_0$ can include any (classical) first-order sentence about $S_0$, such as $h(S_0) > 12$ and $f_1(S_0) \neq 2 \vee f_2(S_0) = 5$.

[2] Free variables in any of these axioms should be understood as universally quantified from the outside.

[3] Entailments are wrt standard Tarskian models, but we will also assume that models assign the usual interpretations to $=$, $<$, $>$, $0$, $1$, $+$, $\times$, $/$, $-$, $e$, $\pi$, and $x^y$ (exponentials).

[4] This captures the idea that the error model of a sensor measuring $f$ depends only on the true value of $f$, and is independent of other factors. In a sense this follows the Bayesian model that conditioning on a random variable $f$ is the same as conditioning on the event of observing $f$. But this is not required in general in the BHL scheme, an issue we ignore for this paper.

[5] Summations can be expressed as logical terms. See BHL.

**Proposition 1:** *Suppose $\mathcal{D}$ includes* (P1)*,* (P2) *and the likelihood axiom for a sensor sense(z) measuring f. Then*

$$\mathcal{D} \models Bel(f = t, do(sense(z), S_0)) = \frac{Bel(f = t, S_0) \cdot Err(z, t)}{\sum_x Bel(f = x, S_0) \cdot Err(z, x)}$$

Essentially, if the robot's sensors are informative, in the sense of returning values closer to the true value, beliefs are strengthened over time.

**From sums to integrals**

While the definition of belief in BHL has many desirable properties, it is defined in terms of a *summation* over situations, and therefore precludes fluents whose values range over the reals. The continuous analogue of (B) then requires *integrating* over some suitable space of values.

As it turns out, a suitable space can be found. First, assume that there are $n$ fluents $f_1, \ldots, f_n$ in $\mathcal{L}$, and that these take no arguments other than the situation argument.[6] Next, suppose that that there is exactly one initial situation for every possible value of these fluents (Levesque *et al.*, 1998):

$$[\forall \vec{x} \exists \iota \bigwedge f_i(\iota) = x_i] \land [\forall \iota, \iota'. \bigwedge f_i(\iota) = f_i(\iota') \supset \iota = \iota'] \quad (I)$$

Under these assumptions, it can be shown that the summation over all situations in (B) can be recast as a summation over all possible initial values $x_1, \ldots, x_n$ for the fluents:

$$Bel(\phi, s) \doteq \frac{1}{\gamma} \sum_{\vec{x}} P(\vec{x}, \phi, s) \quad (B')$$

where $P(\vec{x}, \phi, s)$ is the (unnormalized) weight accorded to the *successor* of an initial world where $f_i$ equals $x_i$:

$$P(\vec{x}, \phi, do(\alpha, S_0)) \doteq$$
$$\text{If } \exists \iota. \bigwedge f_i(\iota) = x_i \land \phi[do(\alpha, \iota)]$$
$$\text{Then} \quad p(do(\alpha, \iota), do(\alpha, S_0))$$
$$\text{Else} \quad 0$$

for any action sequence $\alpha$. In a nutshell, because every situation has an initial situation as an ancestor, and because there is a bijection between initial situations and possible fluent values, it is sufficient to sum over fluent values to obtain the belief even for non-initial situations. Note that unlike (B), this one expects the final situation term $do(\alpha, S_0)$ mentioning what actions and observations took place to be explicitly specified, but that is just what one expects when the agent reasons about its belief after acting and sensing.

The generalization to the continuous case then proceeds as follows. First, we observe that some (though possibly not all) fluents will be real-valued, and that $p(s', s)$ will now be a measure of *density* not weight. For example, if $h$ is real-valued, we might have the following analogue to (4):

$$p(\iota, S_0) = \text{If } 2 \leq h(\iota) \leq 12 \text{ Then } .1 \text{ Else } 0 \quad (5)$$

---
[6]It might be desirable to have fluents take arguments other than the situation. See (Belle and Levesque, 2013a) for discussions.

which says that the true initial value of $h$ is drawn from a uniform distribution on [2,12]. Similarly, the $P$ term above now measures (unnormalized) density rather than weight.

Now suppose fluents are partitioned into two groups: the first $k$ take their values $x_1, \ldots, x_k$ from $\mathbb{R}$, while the rest take their values $y_{k+1}, \ldots, y_n$ from countable domains, then the *degree of belief* in $\phi$ is an abbreviation for:

$$Bel(\phi, s) \doteq \frac{1}{\gamma} \int_{\vec{x}} \sum_{\vec{y}} P(\vec{x} \cdot \vec{y}, \phi, s) \quad (B^+)$$

That is, the belief in $\phi$ is obtained by ranging over all possible fluent values, and integrating[7] and summing the densities of *successor* situations where $\phi$ holds.[8]

To summarize the formalization, a basic action theory $\mathcal{D}$ henceforth is assumed to additionally include: (a) (P1) and (I) as part of $\mathcal{D}_0$; (b) (P2) as part of $\mathcal{D}$'s successor state axioms, and (c) sensor likelihood axioms.

## 3 REGRESSION FOR DISCRETE DOMAINS

We now investigate a computational mechanism for reasoning about beliefs after a trajectory. In this section, we focus on discrete domains, where a *weight*-based notion of belief would be appropriate. Domains with both discrete and continuous variables are reserved for the next section.

Formally, given a basic action theory $\mathcal{D}$, a sequence of actions $\alpha$, we might want to determine whether a formula $\phi$ *holds* after executing $\alpha$ starting from $S_0$:

$$\mathcal{D} \models \phi[do(\alpha, S_0)] \quad (6)$$

which is called *projection* (Reiter, 2001). When it comes to beliefs, and in particular how that changes after acting and sensing, we might be interested in *calculating* the degrees of belief in $\phi$ after $\alpha$: find a real number $n$ such that

$$\mathcal{D} \models Bel(\phi, do(\alpha, S_0)) = n. \quad (7)$$

The obvious method for answering (6) is to translate both $\mathcal{D}$ and $\phi$ into a predicate logic formula. This approach, however, presents a serious computational problem because belief formulas expand into a large number of sentences using (P2), resulting in an enormous search space with initial and successor situations. The other issue with this approach is that sums (and integrals in the continuous case) reduce to complicated second-order formulas.

---
[7]Like in BHL, where summations are captured as logical terms using second-order quantification, we can use logical formulas to capture a variety of sorts of integrals. See (Belle and Levesque, 2013a). We will henceforth simply suppose that for any term $t$ and variable $x$, $\int_x t$ is a term which evaluates (in the standard calculus sense) to the integral of $t$ between $[-\infty, \infty]$.

[8]We are assuming here that the density function is (Riemann) integrable. If it is not or if $\gamma = 0$ then belief is clearly not defined, nor should it be.

We now introduce a *regression* procedure to simplify both (6) and (7) to queries about $Bel(\phi, S_0)$, over arithmetic expressions, for which standard probabilistic reasoning methods can be applied. For this purpose, in the sequel, *Bel* is treated as a *special syntactic operator* rather than as an abbreviation for other formulas. To see a simple example of the procedure, imagine the robot is interested in the probability of $h = 7$, given (4), after reading 5 from a sonar:

$$Bel(h = 7, do(sonar(5), S_0)) \quad (8)$$

If we are to take the sonar's model to be (3), then (8) should be 0 by Bayesian conditioning because the likelihood of the true value being 7 given an observation of 5 is 0. Regression would reduce the term (8) to one over initial priors:

$$\frac{1}{\gamma} \sum_{x \in \{2,\ldots,11\}} Err(5, x) \times Bel(h = x \wedge h = 7, S_0) \quad (9)$$

where *Err* is the error model from (3). By the condition inside *Bel*, the only valid value for $x$ is 7 for which the prior is .1 but $Err(5, 7)$ is 0. Thus, (8) = (9) = 0. In general, regression is a *recursive* procedure that works iteratively over a sequence of actions discarding one action at a time, and it can be utilized to measure any logical property about the variables, e.g. $2\pi \cdot h < 12, h/fuel \leq mileage$, etc.

Formally, regression operates at two levels. (Note that this differs slightly from (Reiter, 2001; Scherl and Levesque, 2003).) At the formula level,[9] we introduce an operator $\mathcal{R}$ for regressing formulas, which over equality literals sends the individual terms to an operator $\mathcal{T}$ for regressing terms. The fundamental objective of these operators is eliminate *do* symbols. The end result, then, is to transform any expression whose situation term is a successor of $S_0$, say $do([a_1, a_2], S_0)$, to one about $S_0$ only, at which point $\mathcal{D}_0$ is all that is needed. As hinted earlier, these operators treat $Bel(\phi, s)$ as though they are special sorts of terms.[10] Throughout the presentation, we assume that the inputs to these operators do not quantify over all situations.

**Definition 2:** For any term $t$, we inductively define $\mathcal{T}[t]$:

1. If $t$ is situation-independent (e.g. $x, \pi^{2/3}$) then $\mathcal{T}[t] = t$.

2. $\mathcal{T}[g(t_1, \ldots, t_k)] = g(\mathcal{T}[t_1], \ldots, \mathcal{T}[t_k])$,
   where $g$ is any non-fluent function (e.g. $\times, +, \mathcal{N}$).

---

[9]For simplicity, in what follows, functional fluents in formulas are only allowed to occur as arguments of an equality literal. It is easy to show that every sentence can be transformed into an equivalent one in the required form, and the transformation is linear in the size of the original sentence, e.g. $h \leq 9$ is written as $\exists u \, (h = u \wedge u \leq 9)$.

[10]In the context of *Bel*, $\phi$ is understood to be any situation-suppressed formula not mentioning $p, l$ and *Bel*. If situation terms do appear in $\phi$, then they may only be the distinguished variable *now*.

3. For a fluent function $f$, $\mathcal{T}[f(s)]$ is defined inductively

   (a) if $s$ is of the form $do(A(\vec{t}), s')$ then
   $\mathcal{T}[f(s)] = \mathcal{T}[e(\vec{t}, s')]$
   
   (b) else $\mathcal{T}[f(s)] = f(s)$
   
   where, in (a), an appropriate instance of the *rhs* of the successor state axiom is used, as obtained from (1).

4. $\mathcal{T}[Bel(\phi, s)]$ is defined inductively:

   (a) if $s$ is of the form $do(a, s')$ and $a$ is a physical action, then
   $$\mathcal{T}[Bel(\phi, s)] = \mathcal{T}[Bel(\psi, s')]$$
   where $\psi$ is $Poss(a, now) \supset \mathcal{R}[\phi[do(a, now)]]$.

   (b) if $s$ is of the form $do(a, s')$ and $a$ is a sensing action $sense(z)$ such that $l(sense(z), s) = Err(z, f_i(s))$ is in $\mathcal{D}$ then
   $$\mathcal{T}[Bel(\phi, s)] =$$
   $$\frac{1}{\gamma} \sum_{x_i} Err(z, x_i) \times \mathcal{T}[Bel(\psi, s')]$$
   where $\psi$ is $Poss(a, now) \supset \phi \wedge f_i(now) = x_i$, and $\gamma$ is the normalization factor and is the same expression as the numerator but $\phi$ replaced by *true*.

   (c) else $\mathcal{T}[Bel(\phi, s)] = Bel(\phi, s)$.

**Definition 3:** For any formula $\phi$, we define $\mathcal{R}[\phi]$ inductively:

1. $\mathcal{R}[t_1 = t_2] = (\mathcal{T}[t_1] = \mathcal{T}[t_2])$

2. $\mathcal{R}[G(t_1, \ldots, t_k)] = G(\mathcal{T}[t_1], \ldots, \mathcal{T}[t_k])$
   where $G$ is any non-fluent predicate (e.g. $=, <$).

3. $\mathcal{R}[Poss(A(\vec{t}), s)] = \mathcal{R}[\Pi_A(\vec{t}, s)]$,
   where an appropriate instance of the *rhs* of the precondition axiom replaces the atom (see Section 2).

4. When $\psi$ is a formula, $\mathcal{R}[\neg \psi] = \neg \mathcal{R}[\psi]$,
   $\mathcal{R}[\forall x \psi] = \forall x \mathcal{R}[\psi], \mathcal{R}[\exists x \psi] = \exists x \mathcal{R}[\psi]$.

5. When $\psi_1$ and $\psi_2$ are formulas,
   $\mathcal{R}[\psi_1 \wedge \psi_2] = \mathcal{R}[\psi_1] \wedge \mathcal{R}[\psi_2]$,
   $\mathcal{R}[\psi_1 \vee \psi_2] = \mathcal{R}[\psi_1] \vee \mathcal{R}[\psi_2]$.

This completes the definition of $\mathcal{T}$ and $\mathcal{R}$. We now go over the justifications for the items, starting with the operator $\mathcal{T}$. In item 1, non-fluents simply do not change after actions. In item 2, $\mathcal{T}$ operates over sums and products in a modular manner. In item 3, provided there are remaining *do* symbols, the physics of the domain determines what the conditions must have been in the previous situation for the current value to hold. In item 4, if there is a remainder physical action, part (a) says that belief in $\phi$ after actions is simply the prior belief about the regression of $\phi$, contingent on action executability. Part (b) says that the belief about

$\phi$ after observing $z$ for the true value of $f_i$ is the prior belief for all possible values $x_i$ for $f_i$ that agree with $\phi$, times the likelihood of $f_i$ being $x_i$ given $z$. The appropriateness of parts (a) and (b) depend on the fact that physical actions do not have any sensing aspect, while sensing actions do not change the world. Part (c) simply says that $\mathcal{T}$ stops when no *do* symbols appear in $s$. We proceed now with the justifications for $\mathcal{R}$. Over equality atoms, $\mathcal{R}$ separates the terms of the equality and sends them to $\mathcal{T}$. Likewise, over non-fluent predicates. When *Poss* is encountered, preconditions take its place. Finally, $\mathcal{R}$ simplifies over connectives in a straightforward way. The main result for $\mathcal{R}$ regarding projection is:

**Theorem 4:** *Suppose $\mathcal{D}$ is any action theory, $\phi$ any situation-suppressed formula and $\alpha$ any action sequence:*

$$\mathcal{D} \models \phi[do(\alpha, S_0)] \quad \text{iff} \quad \mathcal{D}_0 \cup \mathcal{D}_{una} \models \mathcal{R}[\phi[do(\alpha, S_0)]]$$

*where $\mathcal{D}_{una}$ is the unique name assumption and $\mathcal{R}[\phi[do(\alpha, S_0)]]$ mentions only a single situation term, $S_0$.*

Here, $\mathcal{D}_{una}$ is only needed to simplify action terms (Reiter, 2001) e.g. from $fwd(4) = fwd(z)$, $\mathcal{D}_{una}$ infers $z = 4$. Now when our goal is to explicitly compute the degrees of belief in the sense of (7), we have the following property for $\mathcal{T}$:

**Theorem 5:** *Let $\mathcal{D}$ be as above, $\phi$ any situation-suppressed formula and $\alpha$ any sequence of actions. Then:*

$$\mathcal{D} \models Bel(\phi, do(\alpha, S_0)) = \mathcal{T}[Bel(\phi, do(\alpha, S_0))]$$

*where $\mathcal{T}[Bel(\phi, do(\alpha, S_0))]$ is a term about $S_0$ only.*

Theorem 5 essentially shows how belief about trajectories is computable using beliefs about $S_0$ only. Note that, since the result of $\mathcal{T}$ is a term about $S_0$, no sentence outside of $\mathcal{D} - \mathcal{D}_0$ is needed. We now illustrate regression with examples. Using Theorem 5, we reduce beliefs after actions to initial ones. At the final step, standard probabilistic reasoning is applied to obtain the end values.

**Example 6:** Let $\mathcal{D}$ contain the union of (2), (3) and (4).[11] Then the following equality expressions are entailed by $\mathcal{D}$:

1. $Bel(h = 10 \lor h = 11, S_0) = .2$

    $Bel(h \leq 9, S_0) = .8$

    Terms about $S_0$ are unaffected by $\mathcal{T}$. So this amounts to inferring probabilities using $\mathcal{D}_0$.

2. $Bel(h = 11, do(fwd(1), S_0))$

    $= \mathcal{T}[\ Bel(h = 11, do(fwd(1), S_0))\ ]$

    $= \mathcal{T}[\ Bel(\ \underline{\mathcal{R}[(h = 11)[do(fwd(1), now)]]}\ , S_0)\ ]$ (i)

    $= \mathcal{T}[Bel(\ \underline{\mathcal{T}[h(do(fwd(1), now))] = \mathcal{T}[11]}\ , S_0)]$ (ii)

    $= \mathcal{T}[\underline{Bel(\max(0, h - 1) = 11, S_0)}]$ (iii)

    $= Bel(\max(0, h - 1) = 11, S_0)$ (iv)

    $= 0$

    First, since action preconditions are all true, *Poss* is ignored everywhere. We underline to emphasize the expressions undergoing transformations. We begin always by applying $\mathcal{T}$ to the main term, in this case getting (i), by means of $\mathcal{T}$'s item 4(a). Next, $\mathcal{R}$'s item 1 is applied in (ii). While $\mathcal{T}[11] = 11$ by $\mathcal{T}$'s item 1, for $\mathcal{T}[h(do(fwd(1), now))]$ we use item 3 and (2) to get:

    $$\mathcal{T}[\max(0, h(now) - 1)] = \max(0, h(now) - 1)$$

    which is substituted in (ii) to give (iii). Finally, $\mathcal{T}$'s item 4(c) yields (iv), which is a belief term about $S_0$. Now the only valid value for $h$ in (iv) is 12, but for $h = 12$ the robot has a belief of 0 initially.

3. $Bel(h \leq 5, do(sonar(5), S_0))$

    $= \dfrac{1}{\gamma} \displaystyle\sum_{x \in \{2, \ldots, 11\}} Err(5, x) \times \underline{\mathcal{T}[Bel(h = x \land h \leq 5, S_0)]}$ (i)

    $= \dfrac{1}{\gamma} \displaystyle\sum_{x \in \{2, \ldots, 11\}} Err(5, x) \times Bel(h = x \land h \leq 5, S_0)$ (ii)

    $= \dfrac{1}{\gamma} \left( \dfrac{1}{3} \cdot Bel(h = 4 \land h \leq 5, S_0) \right.$

    $\quad + \dfrac{1}{3} \cdot Bel(h = 5 \land h \leq 5, S_0)$ (iii)

    $\quad \left. + \dfrac{1}{3} \cdot Bel(h = 6 \land h \leq 5, S_0) \right)$

    $= \dfrac{1}{\gamma} \left( \dfrac{1}{3} \cdot Bel(h = 4, S_0) + \dfrac{1}{3} \cdot Bel(h = 5, S_0) \right)$ (iv)

    $= \dfrac{1}{\gamma} \cdot \dfrac{2}{30}$

    $= 2/3$

    where $Err(5, x)$ is the model from (3). First, $\mathcal{T}$'s item 4(b) yields (i), and then item 4(c) yields (ii). Since $Err(5, x)$ is non-zero only for $x \in \{4, 5, 6\}$, (ii) is simplified to (iii) and (iv) resulting in $1/15 \cdot 1/\gamma$. We calculate $\gamma$ as follows:

    $= \displaystyle\sum_{x \in \{2, \ldots, 11\}} Err(5, x) \times \mathcal{T}[Bel(h = x \land true, S_0)]$ (i′)

    $= \displaystyle\sum_{x \in \{2, \ldots, 11\}} Err(5, x) \times Bel(h = x, S_0)$ (ii′)

    $= 3/30.$

## 4 REGRESSION FOR GENERAL DOMAINS

We now generalize regression for domains with discrete and continuous variables, for which a *density*-based notion

---

[11] Initial beliefs can also be specified for $\mathcal{D}_0$ using *Bel*, e.g. (4) can be replaced in $\mathcal{D}_0$ with $Bel(h = u, S_0) = .1$ for $u \in \{2, \ldots, 11\}$.

of belief is appropriate. The main issue is that when formulating posterior beliefs after sensing, something like Definition 2's item 4(b) will not work. This is because over continuous spaces the belief about any individual point is 0. Therefore, we will be unpacking belief in terms of the density function, *i.e.* in terms of $P$. These $P(\vec{x}, \phi, s)$ terms, which will now also be treated as special sorts of syntactic terms, are separately regressed. (Of course, the regression of weight-based belief can be approached on similar lines.) Recall that $P(\vec{x}, \phi, S_0)$ is simply the *density* of an initial world (where $f_i = x_i$) satisfying $\phi$. Formally, term regression $\mathcal{T}$ is defined as follows:

**Definition 7:** For any term $t$, we inductively define:

1, 2 and 3 as before.

4. $\mathcal{T}[P(\vec{x}, \phi, s)]$ is defined inductively:

   (a) if $s$ is of the form $do(a, s')$ and $a$ is a physical action then
   $$\mathcal{T}[P(\vec{x}, \phi, s)] = \mathcal{T}[P(\vec{x}, \psi, s')]$$
   where $\psi$ is $Poss(a, now) \supset \mathcal{R}[\phi[do(a, now)]]$.

   (b) if $s$ is of the form $do(a, s')$ and $a$ is a sensing action $sense(z)$ such that $l(sense(z), s) = Err(z, f_i(s))$ is in $\mathcal{D}$, then:
   $$\mathcal{T}[P(\vec{x}, \phi, s)] = Err(z, x_i) \times \mathcal{T}[P(\vec{x}, \psi, s')]$$
   where $\psi$ is $Poss(a, now) \supset \phi \wedge f_i(now) = x_i$.

   (c) else $\mathcal{T}[P(\vec{x}, \phi, s)] = P(\vec{x}, \phi, s)$.

5. $\mathcal{T}[Bel(\phi, s)] = \frac{1}{\gamma} \int_{\vec{z}} \sum_{\vec{y}} \mathcal{T}[P(\vec{z} \cdot \vec{y}, \phi, s)]$.

$\mathcal{R}$ is defined as before. We have the following property:

**Theorem 8:** Let $\mathcal{D}$ be any action theory, $\phi$ any situation-suppressed formula and $\alpha$ any action sequence. Then
$$\mathcal{D} \models Bel(\phi, do(\alpha, S_0)) = \mathcal{T}[Bel(\phi, do(\alpha, S_0))]$$
where $\mathcal{T}[Bel(\phi, do(\alpha, S_0))]$ is a term about $S_0$ only.

Similarly, the analogue of Theorem 4 holds as well.

**Example 9:** Consider the following *continuous* variant of the robot example. Imagine a continuous uniform distribution for the true value of $h$, as provided by (5). Suppose the sonar has the following error profile:

$$l(sonar(z), s) = \text{IF } z \geq 0$$
$$\text{THEN } \mathcal{N}(z - h(s); 0, 4) \quad (10)$$
$$\text{ELSE } 0$$

which says the difference between a nonnegative reading and the true value is normally distributed with mean 0 and variance 4. (A mean of 0 implies there is no systematic bias.) Now, let $\mathcal{D}$ be any action theory that includes (2), (5) and (10). Then the following equalities are entailed by $\mathcal{D}$:

1. $Bel(h = 3 \vee h = 4, S_0) = 0$,
   $Bel(4 \leq h \leq 6, S_0) = .2$

   $\mathcal{T}$ does not change terms about $S_0$. Here, for example, the second belief term equals $\int_4^6 .1 dx = .2$.

2. $Bel(h \geq 11, do(fwd(1), S_0))$
   $$= \frac{1}{\gamma} \int_{x \in \mathbb{R}} \mathcal{T}[\, P(x, \underline{h \geq 11}, do(fwd(1), S_0))\,] \quad \text{(i)}$$
   $$= \frac{1}{\gamma} \int_{x \in \mathbb{R}} \mathcal{T}[P(x, \mathcal{R}[\psi], S_0)] \quad \text{(ii)}$$
   where $\psi$ is $(h \geq 11)[do(fwd(1), now)]$
   $$= \frac{1}{\gamma} \int_{x \in \mathbb{R}} \mathcal{T}[P(x, \underline{\max(0, h-1) \geq 11}, S_0)] \quad \text{(iii)}$$
   $$= \frac{1}{\gamma} \int_{x \in \mathbb{R}} P(x, \max(0, h-1) \geq 11, S_0) \quad \text{(iv)}$$
   $$= \frac{1}{\gamma} \int_{x \in \mathbb{R}} \begin{cases} p(\iota, S_0) & \text{if } \exists \iota.\ h(\iota) = x \wedge h(\iota) \geq 12 \\ 0 & \text{otherwise} \end{cases} \quad \text{(v)}$$
   $$= \frac{1}{\gamma} \int_{x \in \mathbb{R}} \begin{cases} .1 & \text{if } x \in [2, 12] \text{ and } x \geq 12 \\ 0 & \text{otherwise} \end{cases} \quad \text{(vi)}$$
   $$= \frac{1}{\gamma} \int_{x \in \mathbb{R}} \begin{cases} .1 & \text{if } x = 12 \\ 0 & \text{otherwise} \end{cases} \quad \text{(vii)}$$
   $$= 0$$

   We use $\mathcal{T}$'s item 5 to get (i), after which item 4(a) is applied. On doing $\mathcal{R}$ in (ii), along the lines of Example 6.2, we obtain (iii). $\mathcal{T}$'s item 4(c) then yields (iv), and stops. In the steps following (iv), we show how $P$ expands in terms of $p$, and how the space of situations resolves into a mathematical expression, yielding 0.[12]

3. $Bel(h = 0, do(fwd(4), S_0))$
   $$= \frac{1}{\gamma} \int_{x \in \mathbb{R}} \mathcal{T}[P(x, \mathcal{R}[\underline{(h = 0)[do(fwd(4), now)]}], S_0)] \text{ (i)}$$
   $$= \frac{1}{\gamma} \int_{x \in \mathbb{R}} \mathcal{T}[P(x, \max(0, h - 4) = 0, S_0)] \quad \text{(ii)}$$
   $$= \frac{1}{\gamma} \int_{x \in \mathbb{R}} \begin{cases} .1 & \text{if } x \in [2, 12] \text{ and } x \leq 4 \\ 0 & \text{otherwise} \end{cases} \quad \text{(iii)}$$
   $$= .2$$

   By means of (2), after moving forward by 4 units the belief about $h$ is characterized by a *mixed* distribution because $h = 0$ is accorded a .2 weight (*i.e.* from all points where $h \in [2, 4]$ initially), while $h \in (0, 8]$ are associated with a density of .1. Here, $\mathcal{T}$'s item 5 and 4(a) are triggered, and the removal of $\mathcal{T}$ using 4(c) is not shown. The end result is that the density function is integrated for $2 \leq x \leq 4$ leading to .2. ($\gamma$ is 1.)

---

[12]Given certain assumptions, it is possible to further reduce logical expressions involving fluents to a mathematical expression using only those variables that appear in the integral. We expand on this in a longer version of the paper.

4. $Bel(h = 4, do(fwd(-4), do(fwd(4), S_0)))$

$$= \frac{1}{\gamma} \int_{x \in \mathbb{R}} \mathcal{T}[P(x, \exists u. \ h = u \land$$
$$4 = \max(0, u + 4), do(fwd(4), S_0))] \quad \text{(i)}$$

$$= \frac{1}{\gamma} \int_{x \in \mathbb{R}} \mathcal{T}[P(x, \exists u. \ u = \max(0, h - 4) \land$$
$$4 = \max(0, u + 4), S_0)] \quad \text{(ii)}$$

$$= \frac{1}{\gamma} \int_{x \in \mathbb{R}} \begin{cases} .1 & \text{if } x \in [2, 12], \ x \leq 4 \\ 0 & \text{otherwise} \end{cases} \quad \text{(iii)}$$

$$= .2$$

We noted above that the point $h = 4$ gets a .2 weight on executing $fwd(4)$, after which it obtains a $h$ value of 0. The weight is *retained* on reversing by 4 units, with the point now obtaining a $h$ value of 4. The derivation invokes two applications of $\mathcal{T}$'s item 4(a). We skip the intermediate $\mathcal{R}$ steps. ($\gamma$ evaluates to 1.)

5. $Bel(h = 4, do(fwd(4), do(fwd(-4), S_0)))$

$$= \frac{1}{\gamma} \int_{x \in \mathbb{R}} \mathcal{T}[P(x, \exists u. \ u = \max(0, h + 4) \land$$
$$4 = \max(0, u - 4), S_0)] \quad \text{(i)}$$

$$= 0$$

Had the robot moved away first, no "collapsing" of points takes place, $h$ remains a continuous distribution and no point is accorded a non-zero weight. $\mathcal{T}$ steps are skipped but they are symmetric to the one above, *e.g.* compare (i) here and (ii) above. But then the density function is non-zero only for the individual $h = 4$.

6. $Bel(4 \leq h \leq 6, do(sonar(5), S_0))$

$$= \frac{1}{\gamma} \int_{x \in \mathbb{R}} \mathcal{N}(5 - x; 0, 4) \times \mathcal{T}[P(x, \psi, S_0)] \quad \text{(i)}$$
where $\psi$ is $h = x \land 4 \leq h \leq 6$

$$= \frac{1}{\gamma} \int_{x \in \mathbb{R}} \begin{cases} .1 \cdot \mathcal{N}(5 - x; 0, 4) & \text{if } x \in [2, 12], \ x \in [4, 6] \\ 0 & \text{otherwise} \end{cases}$$

$$\approx .41$$

We obtain (i) after $\mathcal{T}$'s item 5 and then 4(b) for sensing actions. That is, belief about $h \in [4, 6]$ is sharpened after observing 5. Basically, we are integrating a function that is 0 everywhere except when $4 \leq x \leq 6$ where it is $.1 \times \mathcal{N}(5 - x; 0, 4)$, normalized over $2 \leq x \leq 12$.

7. $Bel(4 \leq h \leq 6, do(sonar(5), do(sonar(5), S_0)))$

$$= \frac{1}{\gamma} \int_{x \in \mathbb{R}} \mathcal{N}(5 - x; 0, 4) \times \underline{\mathcal{T}[P(x, \psi, s)]} \quad \text{(i)}$$
where $s = do(sonar(5), S_0)$, $\psi$ is $h = x \land 4 \leq h \leq 6$

$$= \frac{1}{\gamma} \int_{x \in \mathbb{R}} [\mathcal{N}(5 - x; 0, 4)]^2 \times \mathcal{T}[P(x, \psi, S_0)] \quad \text{(ii)}$$

$$\approx .52$$

As expected, two successive observations of 5 sharpens belief further. Derivations (i) and (ii) follow from $\mathcal{T}$'s item 5, and two successive applications of item 4(b). Thus, we are to integrate $.1 \times [\mathcal{N}(5 - x; 0, 4)]^2$ between $[4, 6]$ and normalize over $[2, 12]$. These changing densities are plotted in Figure 2.

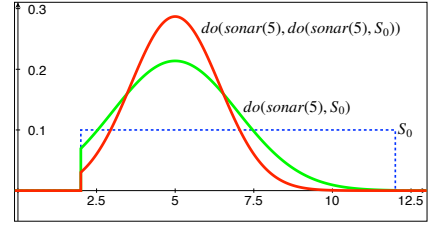

Figure 2: Belief density change for $h$ at $S_0$ (in blue), after sensing 5 (in green) and after sensing 5 twice (in red).

## 5 TWO SPECIAL CASES

Regression is a general property for computing properties about posteriors in terms of priors after actions. It is therefore possible to explore limited cases, which might be appropriate for some applications. We present two such cases.

**Conjugate distributions**

Certain types of systems, such as Gaussian processes, admit an effective propagation model. The same advantages can be observed in our framework. We illustrate this using an example. Assume a fluent $f$, and suppose $\mathcal{D}_0$ is the union of (I), (P1) and the following specification:

$$p(\iota, S_0) = \mathcal{N}(f(\iota); \mu_1, \sigma_1^2)$$

which stipulates that the true value of $f$ is believed to be normally distributed. Assume the following sensor in $\mathcal{D}$:

$$l(sense(z), s) = \mathcal{N}(z - f(s); \mu_2, \sigma_2^2)$$

Then it is easy to show that estimating posteriors yields a product of Gaussians (that is also a Gaussian process (Box and Tiao, 1973)), which is inferred by $\mathcal{T}$:[13]

$$\mathcal{T}[Bel(b \leq f \leq c, do(sense(z), S_0))] =$$
$$\frac{1}{\gamma} \int_b^c \mathcal{N}(x; \mu_1, \sigma_1^2) \cdot \mathcal{N}(z - x; \mu_2, \sigma_2^2) dx$$

**Distribution transformations**

Certain actions affect priors in a characteristically simple manner, and regression would account for these changes as an appropriate function of the initial belief state. We illustrate two instances using Example 9. First, consider an action *grasp(z)* that grabs object *z*. Because the action of grasping does not affect *h* by way of (2), we get:

$$\mathcal{T}[Bel(h \leq b, do(grasp(obj5), S_0))] = Bel(h \leq b, S_0)$$

---
[13]This corresponds to a simple case of Kalman filtering (Dean and Wellman, 1991), where the sensed value is static. In the complete framework with noisy effectors, we would obtain a model where distinct actions may condition priors in distinct ways.

So no changes to *h*'s density are required. Second, consider ground actions with the property that two distinct values of *f* do not become the same after that action, *e.g.*, for initial states this means:

$$\forall \iota, \iota'.\ f(\iota) \neq f(\iota') \supset f(do(a, \iota)) \neq f(do(a, \iota')) \quad \text{(EQ)}$$

Think of *fwd*(−4) that agrees with this, but *fwd*(4) need not. We can show that such actions "shift" priors:

$$\mathcal{T}[Bel(h \leq b, do(fwd(-n), S_0))] = Bel(h \leq b - n, S_0)$$

Intuitively, the probability of *h* being in the interval [*b*, *c*], irrespective of the distribution family, is the same as the probability of $h \in [b+n, c+n]$ after *fwd*(−*n*). Thus, regression derives the initial interval given the current one.[14]

## 6 RELATED WORK

Perhaps the most popular models to treat sensor fusion include variants of Kalman filtering (Fox *et al.*, 2003; Thrun *et al.*, 2005), where priors and likelihoods are assumed to be Gaussian. We already pointed out some instances of Kalman filtering in our example. Where we differ is that backward chaining is possible even when: (a) no assumptions about the nature of distributions, nor about how distributions and dependencies change need to be made, (b) the framework is embedded in a rich theory of actions, and (c) arbitrary forms of incomplete knowledge are allowed, including strict uncertainty.[15] Domain-specific dependencies, then, may be exploited as appropriate.

There have been, of course, other attempts to extend the situation calculus to reason about probabilistic belief, such as (Poole, 1998). See BHL for a discussion on the differences to that work. On a related note, there are numerous approaches that combine logic and probability. In particular, we mention dynamic logic proposals (Van Benthem *et al.*, 2009), planning languages (Younes and Littman, 2004; Sanner, 2011; Kushmerick *et al.*, 1995), and first-order frameworks based on the situation calculus and close relatives (Thielscher, 2001). For discussions on these and non-dynamic proposals such as Markov logics (Richardson and Domingos, 2006), see (Belle and Levesque, 2013a,b).

---

[14]Although *fwd* shifts the distribution linearly, in general, provided something like (EQ) is true, actions may also result in non-linear changes to state variables, which would nonlinearly change the mean of the distribution. Nevertheless, similar features would be observed. See the longer version of the paper for more details.

[15]Moreover, the BHL model is compatible with a wide variety of formalisms such as (Fagin and Halpern, 1994; Halpern and Tuttle, 1993; Halpern, 1990). See BHL for discussions.

There is one other thread of related work, that of *symbolic dynamic programming* (Boutilier *et al.*, 2000, 2001) which also has recent continuous extensions (Sanner *et al.*, 2011). While regression is used in this literature as well, the concerns are very different: they focus on policy generation, while ours is strictly about *belief change*. Consequently, the regression in that literature is adapted from the regression for the non-epistemic situation calculus (Reiter, 2001). Ours, on the other hand, continues in the tradition of the epistemic situation calculus (Scherl and Levesque, 2003) by extending those intuitions to probabilistic belief and noisy sensing. In this regard, our account allows the modeler to explicitly reason about beliefs in the language, which would prove useful in formalizing the achievability of plans (Levesque, 1996), among other things. The idea of regression is not new and lies at the heart of many planning systems (Fritz and McIlraith, 2007). For STRIPS actions, regression has at most linear complexity in the length of the action sequence (Reiter, 2001). For other studies, see (Van Ditmarsch *et al.*, 2007; Rintanen, 2008).

## 7 CONCLUSIONS

Planning and robotic applications have to deal with numerous sources of complexity regarding action and change. Consequently, irrespective of the decompositions and factorizations that are justifiable initially, belief state evolution is known to invalidate these efforts even over simple temporal phenomena. In this paper, we obtain a general methodology to relate beliefs after acting and sensing to initial beliefs. We investigated the methodology in an existing model by BHL, and a continuous extension to it, making the technique applicable to discrete domains as well as general ones. We demonstrated regression using an example where actions affect priors in nonstandard ways, such as transforming a continuous distribution to a mixed one. In general, regression does not insist on (but allows) restrictions to actions, that is, no assumptions need to be made about how actions affect variables and their dependencies over time. Moreover, at the specification level, we do not assume (but allow) structurally constrained initial states.

There are many avenues for future work. Extending automated regression solutions (Reiter, 2001) to subjective probabilities is ongoing work. Moreover, given the promising advances made in the area of relational probabilistic inference, we believe regression suggests natural ways to apply those developments with actions. This line of research would allow us to address effective belief propagation for numerous planning problems that require both logical and probabilistic representations. On another front, note that after applying the reductions, one may also use approximate inference methods. Perhaps then, regression can serve as a computational framework to study approximate belief propagation, on the one hand, and using approximate inference at the initial state after goal regression, on the other.